\documentclass[final]{cvpr}

\usepackage{times}
\usepackage{epsfig}
\usepackage{graphicx}
\usepackage{amsmath}
\usepackage{amssymb}
\usepackage{graphicx}
\usepackage{wrapfig}
\usepackage{booktabs}
\newcommand{\argmin}{\arg\!\min}
\newcommand{\argmax}{\arg\!\max}

\usepackage[pagebackref=true,breaklinks=true,colorlinks,bookmarks=false]{hyperref}

\def\cvprPaperID{1731} 
\def\confYear{CVPR 2021}

\begin{document}

\title{Cyclic Co-Learning of Sounding Object Visual Grounding and Sound Separation}

\author{Yapeng Tian$^{1}$, Di Hu$^{2,3}$$^{\ast}$, Chenliang Xu$^{1}$$^{\ast}$\\
$^{1}$University of Rochester, $^{2}$Gaoling School of Artificial Intelligence, Renmin University of China \\ $^{3}$Beijing Key Laboratory of Big Data Management and Analysis Methods \\
{\tt\small \{yapengtian,chenliang.xu\}@rochester.edu, {dihu@ruc.edu.cn}}
}

\maketitle

\begin{abstract}
There{\let\thefootnote\relax\footnote{{$^{*}$Corresponding authors.}}} are rich synchronized audio and visual events in our daily life. Inside the events, audio scenes are associated with the corresponding visual objects; meanwhile, sounding objects can indicate and help to separate their individual sounds in the audio track. Based on this observation, in this paper, we propose a cyclic co-learning (CCoL) paradigm that can jointly learn sounding object visual grounding and audio-visual sound separation in a unified framework.
Concretely, we can leverage grounded object-sound relations to improve the results of sound separation. Meanwhile, benefiting from discriminative information from separated sounds, we improve training example sampling for sounding object grounding, which builds a co-learning cycle for the two tasks and makes them mutually beneficial. Extensive experiments show that the proposed framework outperforms the compared recent approaches on both tasks, and they can benefit from each other with our cyclic co-learning. {{The source code and pre-trained models are released in \small\url{https://github.com/YapengTian/CCOL-CVPR21}}}.
\end{abstract}

\section{Introduction}
\label{intro}

\begin{figure}
    \centering
    \includegraphics[width=\linewidth]{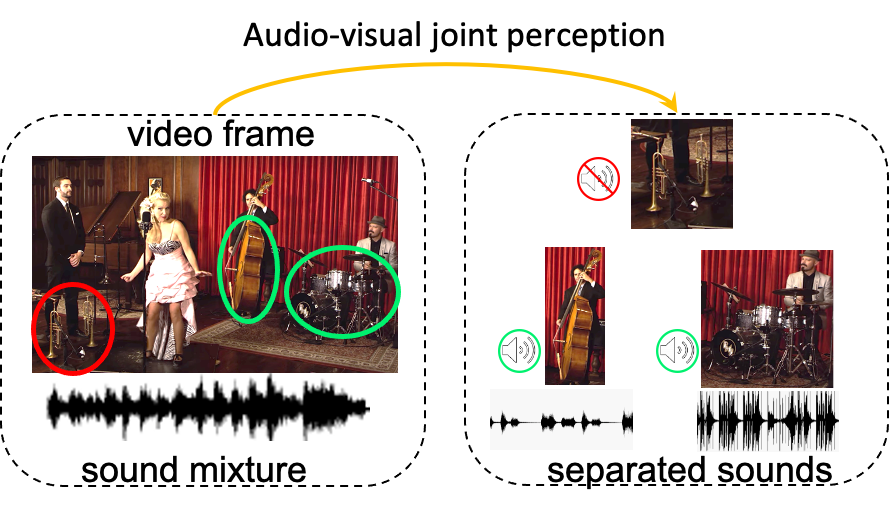}
    \vspace{-5mm}
    \caption{Our model can perform audio-visual joint perception to simultaneously identify silent and sounding objects and separate sounds for individual audible objects.} 
    \label{fig:teaser}
    \vspace{-5mm}
\end{figure}

Seeing and hearing are two of the most important senses for human perception. 
Even though the auditory and visual information may be discrepant, the percept is unified with multisensory integration~\cite{bulkin2006seeing}. Such phenomenon is considered to be derived from the characteristics of specific neural cell, as the researchers in cognitive neuroscience found the superior temporal sulcus in the temporal cortex of the brain can simultaneously response to visual, auditory, and tactile signal~\cite{hikosaka1988polysensory,stein1993merging}.
Accordingly, we tend to perform as unconsciously correlating different sounds and their visual producers, even in a noisy environment.
For example, for a cocktail-party scenario contains multiple sounding and silent instruments as shown in Fig.~\ref{fig:teaser}, we can effortlessly filter out the silent ones and identify different sounding objects, and simultaneously separate the sound for each playing instrument, even faced with a \emph{static visual image}.

For computational models, the multi-modal sound separation and sounding object alignment capacities reflect in audio-visual sound separation (AVSS) and sound source visual grounding (SSVG), respectively. AVSS aims to separate sounds for individual sound sources with help from visual information, and SSVG tries to identify objects that make sounds in visual scenes. These two tasks are primarily explored isolatedly in the literature. Such a disparity to human perception motivates us to address them in a co-learning manner, where we leverage the joint modeling of the two tasks to discover objects that make sounds and separate their corresponding sounds without using annotations. 

Although existing works on AVSS~\cite{ephrat2018looking,owens2018audio,gao2018learning,zhao2018sound,gao2019co,xu2019recursive,gan2020music} and SSVG~\cite{kidron2005pixels,senocak2018learning,tian2018audio,arandjelovic2018objects,qian2020multiple} are abundant, it is non-trivial to jointly learn the two tasks. 
Previous AVSS methods implicitly assume that all objects in video frames make sounds. They learn to directly separate sounds guided by encoded features from either entire video frames~\cite{ephrat2018looking,owens2018audio,zhao2018sound} or detected objects~\cite{gao2019co} without parsing which are sounding or not in unlabelled videos. At the same time, the SSVG methods mostly focus on the simple scenario with single sound source, barely exploring the realistic cocktail-party environment~\cite{senocak2018learning,arandjelovic2018objects}.
Therefore, these methods blindly use information from silent objects to guide separation learning, also blindly use information from sound separation to identify sounding objects.

Toward addressing the drawbacks and enabling the co-learning of both tasks, we introduce a new sounding object-aware sound separation strategy. It targets to separate sounds guided by only sounding objects, where the audiovisual scenario usually consists of multiple sounding and silent objects. To address this challenging task, the SSVG can jump in to help identify each sounding object from a mixture of visual objects, whose objective is unlike previous approaches that make great efforts on improving the localization precision of sound source in simple audiovisual scenario~\cite{senocak2018learning, tian2018audio, arandjelovic2018objects, owens2018audio}.
Accordingly, it is challenging to discriminatively discover isolated sounding objects inside scenarios via the predicted audible regions visualized by heatmaps~\cite{senocak2018learning,arandjelovic2018objects,hu2019deep}. For example, two nearby sounding objects might be grouped together in a heatmap and we have no good principle to extract individual objects from a single located region. 

\begin{figure*}[tb]
    \centering
    \includegraphics[width=\linewidth]{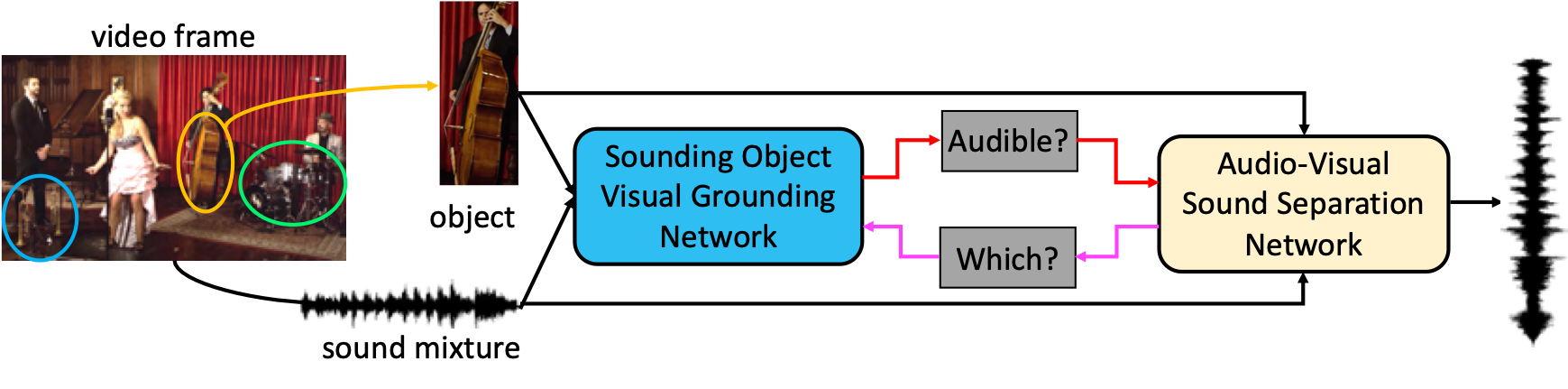}
    \vspace{-6mm}
    \caption{Cyclic Co-learning of sounding object visual grounding and audio-visual sound separation, enabled by sounding object-aware sound separation. Given detected objects from a video and the sound mixture, our model can recognize whether they are audible via a grounding network (\textbf{Audible?}) and separate their sounds with an audio-visual sound separation network to help determine potential sounding and silent sources (\textbf{Which?}).} 
    \label{fig:framework}
    \vspace{-5mm}
\end{figure*}

To enable the co-learning, we propose to directly discover individual sounding objects in visual scenes from visual object candidates. With the help of grounded objects, we learn sounding object-aware sound separation. Clearly, a good grounding model can help to mitigate learning noise from silent objects and improve separation. However, causal relationship between the two tasks cannot ensure separation can further enhance grounding, because they only loosely interacted during sounding object selection. To alleviate the problem, we use separation results to help sample more reliable training examples for grounding. It makes the co-learning in a cycle and both grounding and separation performance will be improved, as illustrated in Fig.~\ref{fig:framework}. Experimental results show that the two tasks can be mutually beneficial with the proposed cyclic co-learning, which leads to noticeable performance and outperforms recent methods on sounding object visual grounding and audio-visual sound separation tasks. 




The main contributions of this paper are as follows: (1) We propose to perform sounding object-aware sound separation with the help of visual grounding task, which essentially analyzes the natural but previously ignored cocktail-party audiovisual scenario. (2) We propose a cyclic co-learning framework 
between AVSS and SSVG to make these two tasks mutually beneficial. (3) Extensive experiments and ablation study validate that our models can outperform recent approaches, and the tasks can benefit from each other with our cyclic co-learning.


\section{Related Work}
\label{relatedwork}

\noindent\textbf{Sound Source Visual Grounding:}
Sound source visual grounding aims to identify the visual objects associating to specific sounds in the daily audiovisual scenario. This task is closely related to the visual localization problem of sound source, which targets to find pixels that are associated with specific sounds. Early works in this field use canonical correlation analysis~\cite{kidron2005pixels} or mutual information~\cite{hershey2000audio,Hu2020CrossTaskTF} to detect visual locations that make the sound in terms of localization or segmentation. Recently, deep audio-visual models are developed to locate sounding pixels based on audio-visual embedding similarities~\cite{arandjelovic2018objects, owens2018audio, hu2019deep, hu2020curriculum}, cross-modal attention mechanisms~\cite{senocak2018learning, tian2018audio,afouras2020self}, vision-to-sound knowledge transfer~\cite{gan2019self}, and sounding class activation mapping~\cite{qian2020multiple}. These learning fashions are capable of predicting audible regions by showing heat-map visualization of single sound source in the simple scenario, but cannot explicitly detect multiple isolated sounding objects when they are sounding at the same time. 
In the most recent work, Hu \emph{et al.}~\cite{hu2020discriminative} propose to discriminatively identify sounding and silent object in the cocktail-party environment, but relying on reliable visual knowledge of objects learned from manually selected single source videos.
Unlike the previous methods, we focus on finding each individual sounding object from cocktail scenario of multiple sounding and silent objects without any human annotations, and is cooperatively learned with audio-visual sound separation task.


\vspace{1mm}
\noindent\textbf{Audio-Visual Sound Separation:} 
Respecting for the long-history research on sound source separation in signal processing, we only survey recent audio-visual sound source separation methods~\cite{ephrat2018looking,owens2018audio,gao2018learning,zhao2018sound,gao2019co,xu2019recursive,zhao2019sound,rouditchenko2019self,gan2020music} in this section. These approaches separate visually indicated sounds for different audio sources (\emph{e.g.}, speech  in~\cite{ephrat2018looking,owens2018audio}, music instruments in~\cite{zhao2018sound,gao2019co,zhao2019sound,gan2020music}, and universal sources  in~\cite{gao2018learning,rouditchenko2019self}) with a commonly used mix-and-separate strategy to build training examples~\cite{hershey2016deep,huang2015joint}. Concretely, they generate the corresponding sounds w.r.t. the given visual objects~\cite{zhao2018sound} or object motions in the video~\cite{zhao2019sound}, where the objects are assumed to be sounding during performing separation.
Hence, if a visual object belonging to the training audio source categories is silent, these models usually fail to separate a all-zero sound for it.
Due to the commonly existing audio spectrogram overlapping phenomenon, these methods will introduce artifacts during training and make wrong predictions during inference. Unlike previous approaches, we propose to perform sounding object-aware audio separation to alleviate the problem. Moreover, sounding object visual grounding is explored in a unified framework with the separation task. 

\vspace{1mm} 
\noindent\textbf{Audio-Visual Video Understanding:} Since auditory modality containing synchronized scenes as the visual modality is widely available in videos, it attracts a lot of interests in recent years.
Besides sound source visual grounding and separation, a range of audio-visual video understanding tasks including 
audio-visual action recognition~\cite{gao2019listentolook,kazakos2019epic,korbar2019scsampler},
audio-visual event localization~\cite{tian2018audio,lin2019dual,wu2019DAM}, audio-visual video parsing~\cite{tian2020avvp}, cross-modal generation~\cite{chen2018lip,chen2019hierarchical,gao20192,zhou2020sep,Zhou2021pose,Xu2021visual}, and audio-visual video captioning~\cite{rahman2019watch,Tian_2019_CVPR_Workshops,wang2018watch} have been explored. Different from these works, we introduce a cyclic co-learning framework for both grounding and separation tasks
and show that they can be mutually beneficial. 


\section{Method}

\subsection{Overview}
\label{sec:CCoL}

 Given an unlabeled video clip $V$ with the synchronized sound $s(t)$\footnote{$s{(t)}$ is a time-discrete audio signal.}, $\mathcal{O} = \{O_1,...,O_N\}$ are $N$ detected objects in the video frames and the sound mixture is $s(t) = \sum_{n=1}^{N}s_n(t)$. Here, $s_n(t)$ is the separated sound of the object $O_n$. When it is silent, $s_n(t) = 0$. Our co-learning aims to recognize each sounding object $O_n$ and then separate its sound $s_n(t)$ for the object.

The framework, as illustrated in Fig.~\ref{fig:framework}, mainly contains two modules: sounding object visual grounding network and audio-visual sound separation network. 
The sounding object visual grounding network can discover isolated sounding objects from object candidates inside video frames. We learn the grounding model from sampled positive and negative audio-visual pairs.  
To learn sound separation in the framework, we adopt a commonly used mix-and-separate strategy~\cite{gao2019co,hershey2016deep,zhao2018sound} during training. Given two training video and sound pairs $\{V^{(1)}, s^{(1)}(t)\}$ and $\{V^{(2)}, s^{(2)}(t)\}$, we obtain a mixed sound: 
\begin{equation}
    s_m(t) = s^{(1)}(t) + s^{(2)}(t) = \sum_{n=1}^{N_{1}}s^{(1)}_n(t) + \sum_{n=1}^{N_{2}}s^{(2)}_n(t),
    \label{wave_relation}
\end{equation}
and find object candidates $\mathcal{O}^{(1)} = \{O_1^{(1)},...,O_{N_1}^{(1)}\}$ and $\mathcal{O}^{(2)} = \{O_1^{(2)},...,O_{N_2}^{(2)}\}$ from the two videos. The sounding object visual grounding network will recognize audible objects from $\mathcal{O}^{(1)}$ and $\mathcal{O}^{(2)}$ and the audio-visual sound separation network will separate sounds for the grounded objects. Sounds are processed in a Time-Frequency space with the short-time Fourier
transform (STFT).

With the sounding object-aware sound separation, we can co-learn the two tasks and improve audio-visual sound separation with the help of sounding object visual grounding. However, the separation performance will highly rely on the grounding model and the grounding task might not benefit from co-learning training due to the weak feedback from separation.
To simultaneously evolve the both models, we propose a cyclic co-learning strategy as illustrated in Fig.~\ref{fig:cycle}, which has an additional backward process that utilizes separation results to directly improve training sample mining for sounding object visual grounding. In this manner, we can make the two tasks mutually beneficial.

\subsection{Sounding Object Visual Grounding}
\label{sec:sovg}

Videos contain various sounds and visual objects, and not all objects are audible.  
To find sounding objects in videos $V^{(1)}$ and $V^{(2)}$ and further utilize grounding results for separation, we formulate sounding object visual grounding as a binary matching problem. 

\noindent \textbf{Sounding Object Candidates:} 
To better support the audio-visual matching problem, we choose to follow the widely-adopted image representation strategy of visual object proposal in image captioning~\cite{karpathy2015deep,anderson2018bottom},
which has been also employed in the recent work on audio-visual learning~\cite{gao2019co}. 
Concretely, the potential audible visual objects are first proposed from videos using an object detector. In our implementation, we use the Faster R-CNN~\cite{ren2015faster} object detector trained on Open Images dataset~\cite{krasin2017openimages} from~\cite{gao2019co} to detect objects from video frames in $V^{(1)}$ and $V^{(2)}$ and obtain $\mathcal{O}^{(1)} = \{O_1^{(1)},...,O_{N_1}^{(1)}\}$ and $\mathcal{O}^{(2)} = \{O_1^{(2)},...,O_{N_2}^{(2)}\}$. Next, we learn to recognize sounding objects in $\mathcal{O}^{(1)}$ and $\mathcal{O}^{(2)}$ associated with $s^{(1)}(t)$ and $s^{(2)}(t)$, respectively. For simplicity, we use an object $O$ and a sound $s(t)$ as an example to illustrate our grounding network.                              

\noindent \textbf{Audio Network:} Raw waveform $s(t)$ is transformed to an audio spectrogram $S$ with the STFT. An VGG~\cite{simonyan2014very}-like 2D CNN network: VGGish followed by a global max pooling (GMP) is used to extract an audio embedding $f_s$ from $S$.   

\noindent \textbf{Visual Network:} The visual network extracts features from detected visual object $O$. We use the pre-trained ResNet-18~\cite{he2016deep} model before the last fully-connected layer to extract a visual feature map and perform a GMP to obtain a visual feature vector $f_o$ for $O$.

\noindent \textbf{Grounding Module:} The audio-visual grounding module takes audio feature $f_s$ and visual object feature $f_o$ as inputs to predict whether the visual object $O$ is one of the sounding makers for $s(t)$. We solve it using a two-class classification network. It first concatenates $f_s$ and $f_o$ and then uses a 3-layer Multi-Layer Perceptron (MLP) with a Softmax function to output a probability score $g(s(t), O) \in \mathcal{R}^2$. Here, if $g(s(t), O)[0] >= 0.5$, $\{s(t), O\}$ is a positive pair and $s(t)$ and $O$ are matched; otherwise, $O$ is not a sound source. 

\noindent \textbf{Training and Inference:} To train the sounding object visual grounding network, we need to sample positive/matched and negative/mismatched audio and visual object pairs. It is straightforward to obtain negative pairs with composing audio and objects from different videos. For example, $s^{(1)}(t)$ from $V^{(1)}$ and an randomly selected object $O_r^{(2)}$ from $V^{(2)}$ can serve as a negative pair. However, positive audio-visual pairs are hard to extract since not all objects are audible in videos. If an object from $V^{(1)}$ is not audio source, the object and $s^{(1)}(t)$ will be a negative pair, even though they are from the same video. To address the problem, we cast the positive sample mining as a multiple instance learning problem and sample the most confident pair as a positive sample with a grounding loss as the measurement: 
\begin{equation}
  \hat{n} =  \argmin_n f(g(s^{(1)}(t), O^{(1)}_n), y_{pos}),
  \label{eq:pos_min}
\end{equation}
where $f(\cdot)$ is a cross-entropy function; $y_{pos} = [1, 0]$ is an one-hot encoding for positive pairs; $O_{\hat{n}}^{(1)}$ and $s^{(1)}$ will be the positive audio-visual pair for training. With the sampled negative and positive data, we can define the loss function to learn the sounding object visual grounding:
\begin{equation}
    l_{grd_{s}} = f(g(s^{(1)}(t), O^{(2)}_r), y_{neg}) +  f(g(s^{(1)}(t), O^{(1)}_{\hat{n}}), y_{pos}),
\end{equation}
where $y_{neg} = [0, 1]$ is the negative label. The visual grounding network can be end-to-end optimized with sampled training pairs via $l_{grd_s}$. 

During inference, we can feed audio-visual pairs $\{O_i^{(1)}, s^{(1)}(t)\}_{i=1}^{N_1}$ and $\{O_i^{(2)}, s^{(2)}(t)\}_{i=1}^{N_2}$ into the trained model to find sounding objects insides the two videos. To facilitate audio-visual sound separation, we need to detect sounding objects from the sound mixture $s_m(t)$ rather than $s^{(1)}(t)$  and $s^{(2)}(t)$, since the individual sounds are unavailable at a testing stage for separation task. 
 
\subsection{Sounding Object-Aware Separation}
\label{sec:viss}

Given detected objects in $\mathcal{O}^{(1)}$ and $\mathcal{O}^{(2)}$, we separate sounds for each object from the sound mixture $s_m(t)$ and mute separated sounds of silent objects. 


Using an audio-visual sound separation network, we can predict sound spectrograms $\{S_n^{(1)}\}_{n=1}^{N_1}$ and $\{S_n^{(2)}\}_{n=1}^{N_2}$ for objects in $\mathcal{O}^{(1)}$ and $\mathcal{O}^{(2)}$, respectively. 
According to waveform relationship in Eq.~\ref{wave_relation}, we can approximate spectrogram magnitude relationship as: $S^{(1)} \approx \sum_{n=1}^{N_{1}}S^{(1)}_n  \quad \textrm{and} \quad S^{(2)} \approx \sum_{n=1}^{N_{2}}S^{(2)}_n.$
To learn the separation network, we can optimize it with a L1 loss function:
\begin{equation}
    l_{sep} =  ||S^{(1)} - \sum_{n=1}^{N_{1}}S^{(1)}_n||_1  + ||S^{(2)} - \sum_{n=1}^{N_{2}}S^{(2)}_n||_1.
    \label{sep_loss_base}
\end{equation}
However, not all objects are audible and spectrograms from different objects contain overlapping content. Therefore, even an object $O^{(1)}_n$ is not sounding, it can also separate non-zero sound spectrogram $S^{(1)}_n$ from the spectrogram of sound mixture $S_m$, which will introduce errors during training. To address the problem, we propose a sounding object-aware separation loss function: 
\begin{multline}
    l_{sep}^{*} =  ||S^{(1)} - \sum_{n=1}^{N_{1}}g^{*}(s_m(t), O^{(1)}_n)S^{(1)}_n||_1 \\ + ||S^{(2)} - \sum_{n=1}^{N_{2}}g^{*}(s_m(t), O^{(2)}_n)S^{(2)}_n||_1, 
\end{multline}
 where $g^{*}(\cdot)$ is a binarized value of $g(\cdot)[0]$. If an object $O^{(1)}_n$ is not a sound source, $g^{*}(s_m(t), O^{(1)}_n)$ will be equal to zero. Thus, the sounding object-aware separation can help to reduce training errors from silent objects in Eq.~\ref{sep_loss_base}. 
 
In addition, we introduce additional grounding loss terms to guide the grounding model learning from the sound mixture. Since we have no sounding object annotations, we adopt a similar positive sample mining strategy as in Eq.~\ref{eq:pos_min} and define a grounding loss as follows:
\begin{align}
    l_{{grd_m}} = \sum_{k=1}^2\min_n f(g(s_m(t), O^{(k)}_n), y_{pos}). 
\end{align}

\subsection{Co-learning in a Cycle}
\label{sec:cycle}
\begin{figure}
\begin{center}
\includegraphics[width=0.4\textwidth]{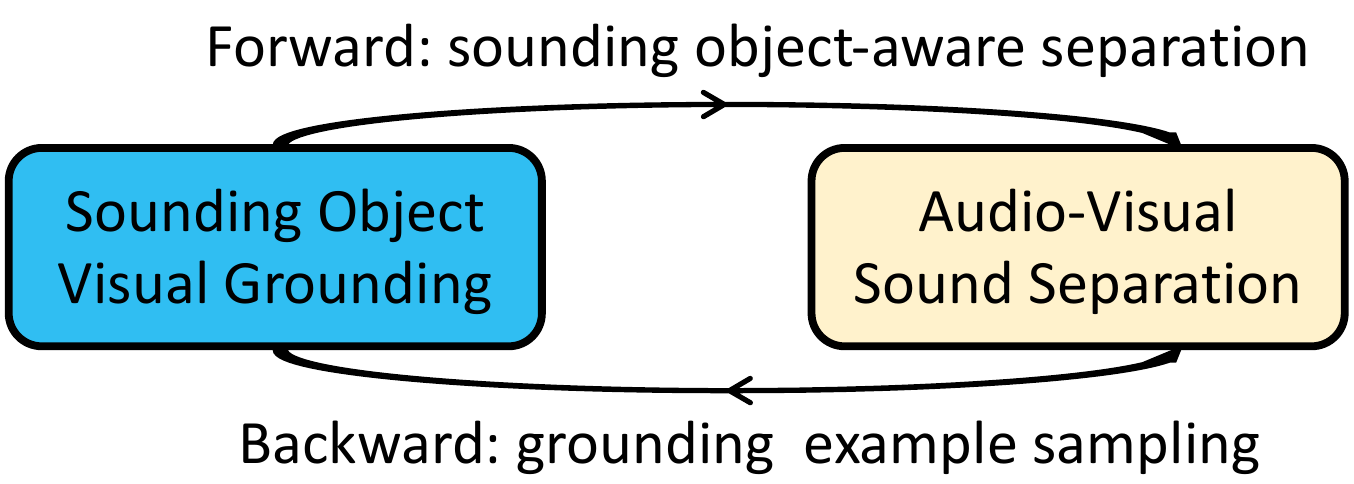}
\end{center}
\vspace{-3mm}
   \caption{Cyclic co-learning. Facilitated by sounding object visual grounding, our model can employ sounding object-aware sound separation to improve separation. Meanwhile, separation results can help to do effective training sample mining for grounding.}
\label{fig:cycle}
\vspace{-6mm}
\end{figure}
Combing grounding and separation loss terms, we can learn the two tasks in a unified way with a co-learning objective function: $l_{col} = l_{grd_s} + l_{sep}^{*} + l_{grd_m}.$

Although our sounding object visual grounding and audio-visual sound separation models can be learned together, the two tasks loosely interact in $l_{sep}^{*}$. Clearly, a good grounding network can help improve the separation task. However, the grounding task might not be able to benefit from co-learning training since there is no strong feedback from separation to guide learning the grounding model. To further strengthen the interaction between the two tasks, we propose a cyclic co-learning strategy, which can make them benefit from each other. 

If an object $O_n^{(k)}$ makes sound in video $V^{(k)}$, the separated spectrogram $S^{(k)}_n$ should be close to $S^{(k)}$; otherwise, the difference between $S^{(k)}_n$ and $S^{(k)}$ should be larger than a separated sound spectrogram from an sounding object and $S^{(k)}$. We use $L_1$ distance to measure dissimilarity of spectrograms: $d^{(k)}_{n} = ||S^{(k)}_n - S^{(k)}||_1$,
where $d^{(k)}_{n}$ will be small for a sounding object $O^{(k)}_{n}$. Based on the observation, we select the object $O_n^{(k)}$ with the minimum $d^{(k)}_{n}$ make the dominant sound in $V^{k}$ to compose positive samples for sounding object visual grounding. Let $\hat{n}_1 =  \argmin_n d^{(1)}_{n}$ and $\hat{n}_2 =  \argmin_n d^{(2)}_{n}$. We can re-formulate grounding loss terms as:
\begin{align*}
    l_{grd_s}^* &=  f(g(s^{(1)}(t), O^{(2)}_r), y_{neg}) +  f(g(s^{(1)}(t), O^{(1)}_{\hat{n}_1}), y_{pos}),\\
    l_{grd_m}^{*} &=f(g(s_m(t), O^{(1)}_{\hat{n}_1}), y_{pos}) +  f(g(s_m(t), O^{(2)}_{\hat{n}_2}), y_{pos}).
\end{align*}

In addition, if $d^{(k)}_{n}$ is very large, the object $O_n^{(k)}$ is very likely not be audible, which can help us sample potential negative examples for mixed sound grounding. Specifically, we select the objects that are associated with the largest $d^{(k)}_{n}$, and $d^{(k)}_{n}$ must be larger than a threshold $\epsilon$. Let $n_1^{*} =  \argmax_n d^{(1)}_{n}$, s.t. $d^{(1)}_{n}>\epsilon $ and $n_2^{*} =  \argmax_n d^{(2)}_{n}$, s.t. $d^{(2)}_{n}>\epsilon$. We can update $l_{grd_m}^*$ with learning from potential negative samples: $\hat{l}_{grd_m}^{*} = \sum_{k=1}^{2} (f(g(s_m(t), O^{(k)}_{\hat{n}_k}), y_{pos}) +  f(g(s_m(t), O^{(k)}_{n_k^*}), y_{neg})).$
Finally, we can co-learn the two tasks in a cycle with optimizing the joint cyclic co-learning loss function: $l_{ccol} = l_{grd_s}^* + l_{sep}^* + \hat{l}_{grd_m}^*$. Inside cyclic co-learning as illustrated in Fig.~\ref{fig:cycle}, we use visual grounding to improve sound separation and enhance visual grounding based on feedback from sound separation. The learning strategy can make the tasks help each other in a cycle and significantly improve performance for both tasks. 

\vspace{-2mm}
\section{Experiments}
\label{exp}

\subsection{Experimental Setting} 
\label{expset}

\noindent \textbf{Dataset:} In our experiments, 520 online available musical solo videos from the widely-used MIT MUSIC dataset~\cite{zhao2018sound} is used. The dataset includes 11 musical instrument categories: accordion, acoustic guitar, cello, clarinet, erhu, 
ute, saxophone, trumpet, tuba, violin, and xylophone. The dataset is relatively clean and sounding instruments are usually visible in videos. We split it into training/validation/testing sets, which have 468/26/26 videos from different categories, respectively. To train and test our cyclic co-learning model, we randomly select three other videos for each video to compose training and testing samples. Let's denote the four videos as A, B, C, D. We compose A, B together as $V^{(1)}$ and C, D together as $V^{(2)}$, while sounds of $V^{(1)}$ and $V^{(2)}$ are only from A and C, respectively. Thus, objects from B and D in the composed samples are inaudible. Finally, we have 18,720/260/260 composed samples in our training/val/test sets for the two tasks.

\noindent \textbf{Evaluation Metrics:} For sounding object visual grounding, we feeding detected audible and silent objects in videos into different grounding models and evaluate their binary classification accuracy. We use the mir eval library~\cite{raffel2014mir_eval} to measure sound separation performance in terms of two standard metrics:
Signal-to-Distortion Ration (SDR) and Signal-to-Interference
Ratio (SIR). 

\begin{table}
  \centering
  \scalebox{0.76}{
  \begin{tabular}{l|cc|ccc}
    \toprule
    Methods     & OTS~\cite{arandjelovic2018objects}    & DMC~\cite{hu2019deep} & Grounding only &CoL &CCoL \\
    \midrule
    Single Sound & 58.7  & 65.3 & \noindent \textbf{72.0}& 67.0&\noindent \underline{\textbf{84.5}}    \\
    Mixed Sound & 51.8  & 52.6 &\noindent \textbf{61.4}&58.2&\noindent \underline{\textbf{75.9}}    \\
    \bottomrule
  \end{tabular}
  }
  \vspace{1mm}
    \caption{Sounding object visual grounding performance (\%). Top-2 results are highlighted.}
  \label{tbl:grd}
  \vspace{-5mm}
\end{table}

\begin{figure*}[ht]
    \centering
    \includegraphics[width=\linewidth]{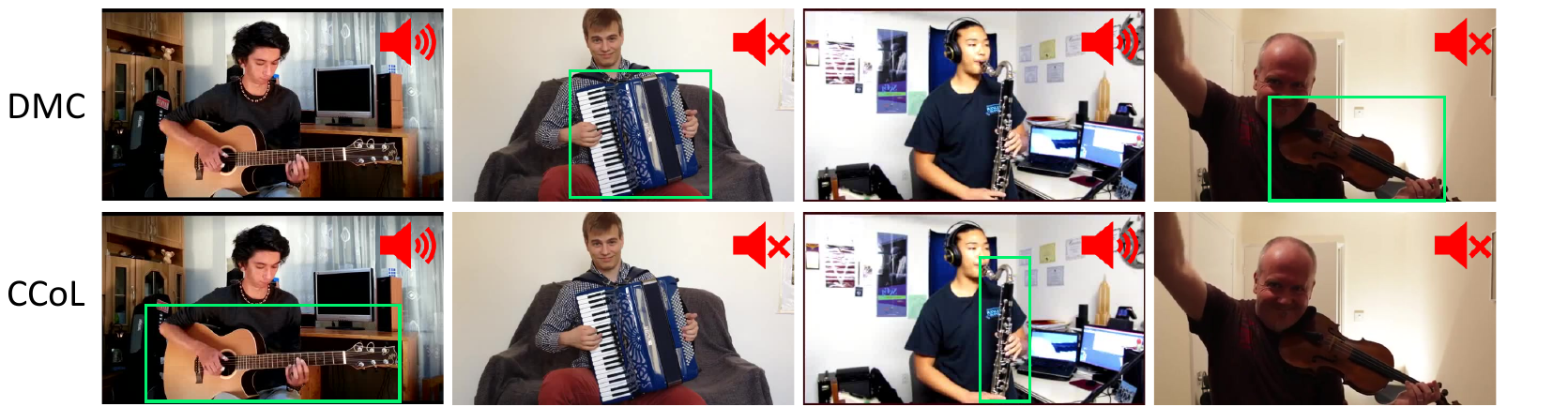}
    \caption{Qualitative results of sounding object visual grounding for both audible and silent objects. We use two icons to denote whether objects in video frames are audible or not and grounded sounding objects from DMC and CCoL are shown in green boxes. Our CCoL model can effectively identify both sounding and silent objects, while the DMC fails. Note that 2-sound mixtures are used as inputs.}
    \label{fig:grd}
    \vspace{-3mm}
\end{figure*}

\begin{figure*}[ht]
    \centering
    \includegraphics[width=0.95\linewidth]{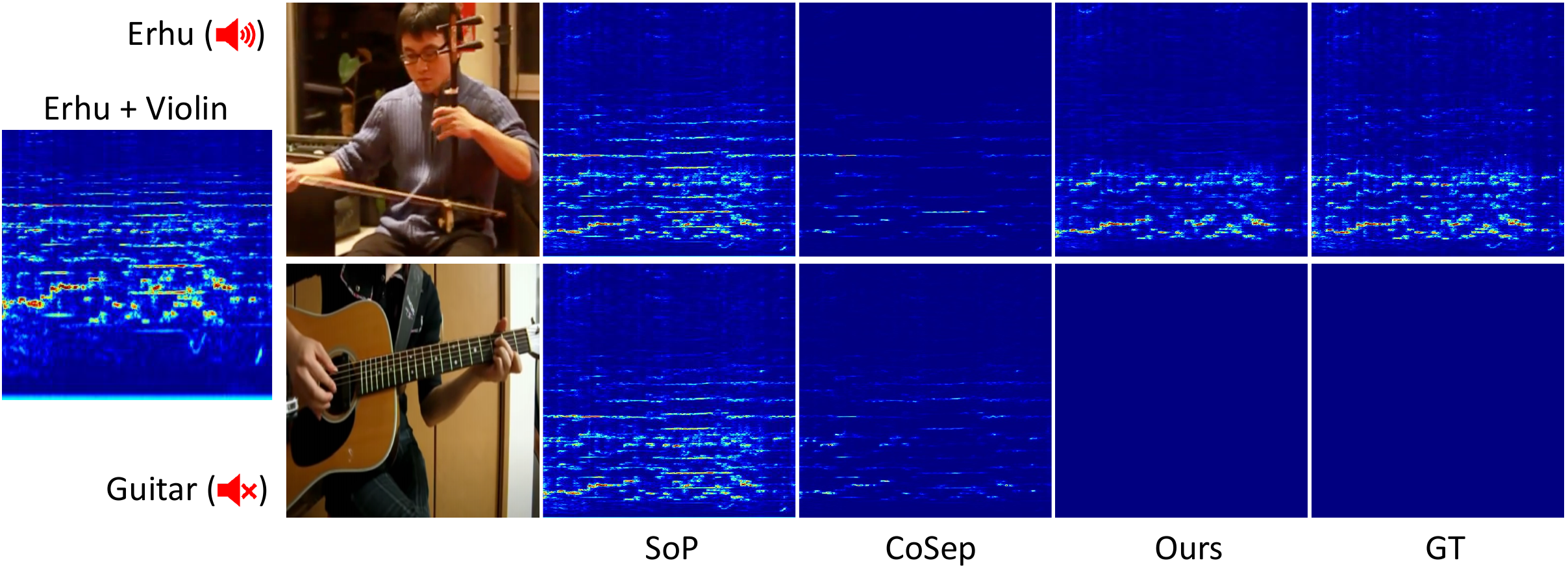}
    \caption{Qualitative results of audio-visual sound separation for both audible and silent objects. Our CCoL model can well mitigate learning noise from silent objects during training and generate more accurate sounds.}
    \label{fig:sep}
    \vspace{-3mm}
\end{figure*}

\noindent \textbf{Implementation Details:}
We sub-sample audio signals at 11kHz, and each video sample is approximately 6 seconds. The STFT is calculated using a Hann window size of 
1022 and a hop length of 256 and each 1D audio waveform is transformed to a $512\times 256$ Time-
Frequency spectrogram. Then, it is re-sampled to $T, F$ = 256. The video frame rate is set as 1$fps$ and we randomly select 3 frames per 6$s$ video. Objects extracted from video frames are resized to $256\times 256$ and then randomly cropped to $224\times 224$ as inputs to our network. $\epsilon$ is set to $0.1$.
We use a soft sound spectrogram masking strategy as in ~\cite{gao2019co, zhao2018sound} to generate individual sounds from audio mixtures and adopt a audio-visual 
sound separation network from~\cite{zhao2018sound}. More details about the network can be found in our appendix.
Our networks are optimized by Adam~\cite{kingma2014adam}. 
Since the sounding object visual grounding and audio-visual sound separation tasks are mutually related, we need to learn good initial models for making them benefit from cyclic co-learning. To this end,
we learn our CCoL model with three steps in a curriculum learning~\cite{bengio2009curriculum} manner. Firstly, we train the sounding object visual grounding network with $l_{grd_s}$. Secondly, we co-learn the grounding network initialized with pre-trained weights and the separation network optimized by $l_{col}$. Thirdly, we use $l_{ccol}$ to further fine-tune the two models.


\begin{figure*}[ht]
    \centering
    \includegraphics[width=\linewidth]{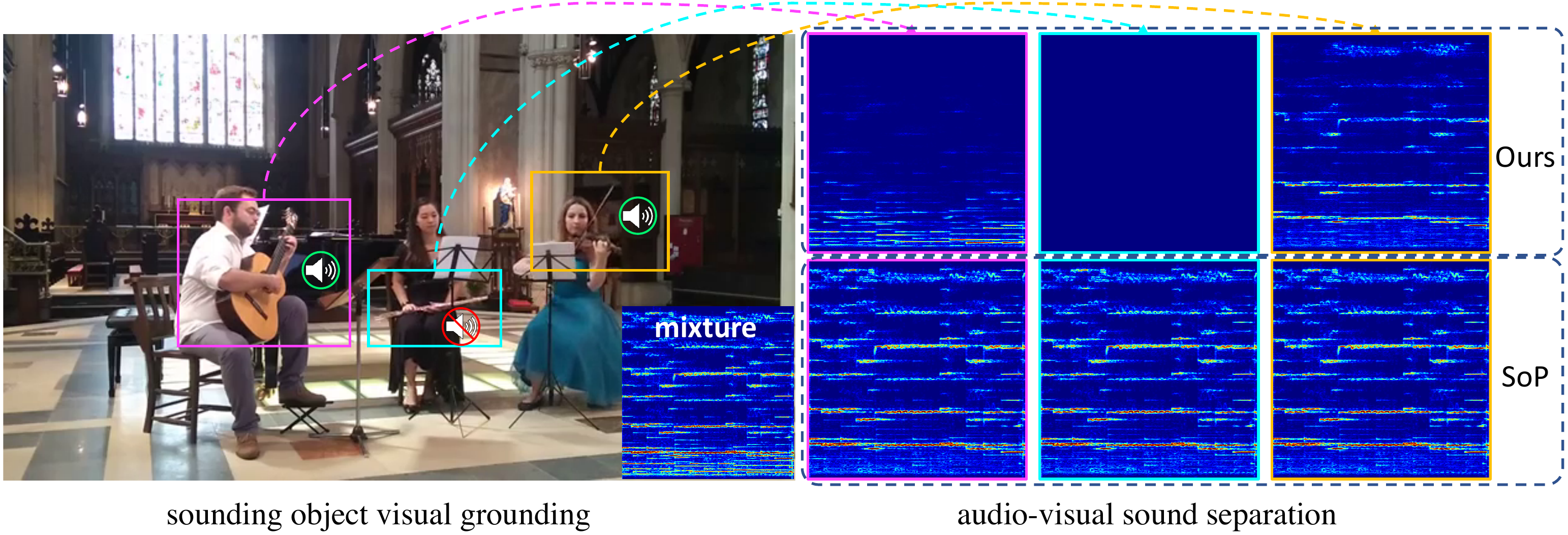}
    \caption{Real-world sounding object visual grounding and audio-visual sound separation. The \textit{guitar} and \textit{violin} are playing and the \textit{flute} is visible but not audible. Our model can identify sounding objects: \textit{guitar} and \textit{violin} and silent object: \textit{flute}. Moreover, it can simultaneously separate individual sounds for each instrument, while the SoP using the same noisy training data as ours fails to associate objects with the corresponding sounds, thus obtains poor separation results.}
    \label{fig:joint_loc_sep}
    \vspace{-3mm}
\end{figure*}

\subsection{Sounding Object Visual Grounding}
 We compare our methods to two recent methods: OTS~\cite{arandjelovic2018objects} and DMC~\cite{hu2019deep}. In addition, we make an ablation study to investigate the proposed models. The Grounding only model is trained only with grounding losses: $l_{grd_s}$; the co-learning (CoL) model jointly learn visual grounding and sound separation using the $l_{col}$; and the cyclic co-learning (CCoL) further strengthens the interaction between the two tasks optimized via $l_{ccol}$. We evaluate sounding object visual grounding performance on both solo and mixed sounds. 
 
Table~\ref{tbl:grd} and Figure~\ref{fig:grd} show quantitative and qualitative sounding object visual grounding results,respectively. Even our grounding only has already outperformed the OTS and DMC, which can validate the effectiveness of the proposed positive sample mining approach. Then, we can see that the CoL with jointly learning grounding and separation achieves worse performance than the Grounding only model. It demonstrates that the weak interaction inside CoL cannot let the grounding task benefit from the separation task. However, using separation results to help the grounding example sampling, our CCoL is significantly superior over both Grounding only and CoL models. The results can demonstrate the sounding object visual grounding can benefit from separation with our cyclic learning.

\begin{table}
  \centering
  \scalebox{0.6}{
  \begin{tabular}{l|ccc|ccc|c}
    \toprule
    Methods  &RPCA~\cite{huang2012singing}   &SoP~\cite{zhao2018sound} & CoSep~\cite{gao2019co} &Random Obj&CoL &CCoL &Oracle\\
    \midrule
    SDR  &-0.48& 3.42 &2.04 &4.20 &\noindent \textbf{6.50} &\noindent \underline{\textbf{7.27}} &7.71\\
     SIR  &3.13 & 4.98 & 6.21 &6.90 &\noindent\textbf{11.81} &\noindent \underline{\textbf{12.77}}  &11.42\\
    \bottomrule
  \end{tabular}
  }
   \caption{Audio-visual sound separation performance. Top-2 results are highlighted.}
   \vspace{-3mm}
  \label{tbl:sep}
\end{table}

\subsection{Audio-Visual Sound Separation}
To demonstrate the effectiveness of our CCoL framework on audio-visual sound separation, we compare it to a classical factorization-based method: RPCA~\cite{huang2012singing} and two recent state-of-the-art methods: SoP~\cite{zhao2018sound}, CoSep~\cite{gao2019co}, and baselines: Random Obj and CoL in Tab.~\ref{tbl:sep}. The SoM~\cite{zhao2019sound} and Music Gesture~\cite{gan2020music} address music sound separation by incorporating dynamic visual motions, and show promising results. However, as the SoP and CoSep, they also did not consider the silent object problem. Meanwhile, since there are no source code for the two appraoches, we will not include them into comparison. \textit{Note that SoP and CoSep are trained using source code provided by the authors and the same training data (including audio preprocessing) as ours}. Moreover, we show separation results of an Oracle, which feeds ground truth grounding labels of mixed sounds to train the audio-visual separation network.  

We can see that our CoL outperforms the compared SoP, CoSep, and Random Obj, and CCoL is better than CoL. The results demonstrate that sounding object visual grounding in the co-learning can help to mitigate training errors from silent video objects in separation, and separation performance can be further improved with the help of enhanced grounding model by cyclic co-learning. Compared to the Oracle model, it is reasonable to see that CCoL has slightly lower SDR. A surprising observation is that CoL and CCoL achieve better results in terms of SIR. One possible reason is that our separation networks can explore various visual objects as inputs during joint grounding and separation learning, which might make the models more robust on SIR.

Moreover, we illustrate quantitatively results for testing videos with both audible and silent objects in Tab.~\ref{tbl:sep_silent}. Both SoP and CoSep are blind to whether a visual object makes sound and they will generate non-zero audio waveform for silent objects, and the Random Obj is limited in identifying object silent and sounding objects, thus they will have even lower SDRs and SIRs. However, both CoL and CCoL are capable of recognizing the audibility of objects and employ sounding object-aware separation, which helps the two models achieve significant better results. The experimental results can well validate the superiority of the proposed sounding object-aware separation mechanism. 

We further show qualitative separation results for audible and silent objects in Fig.~\ref{fig:sep}. We can see that both SoP and CoSep generate nonzero spectrograms for the silent \textit{Guitar} and our CCoL can separate much better \textit{Erhu} sound. The results can validate that the CCoL model is more robust to learning noise from silent objects during training and can effectively perform sounding object-aware sound separation. 

Moreover, we train our model by letting the video 'B' and 'D' be randomly chosen from “with audio” and “silent”. In this way, each training video contains one or two sounding objects and the corresponding mixed sounds will be from up to four different sources. 
The SDR/SIR scores from SoP and CoL, and CCoL are 2.73/4.08, 5.79/11.43, and 6.46/11.72, respectively. The results further validate that the proposed cyclic co-learning framework can learn from both solo and duet videos and is superior over naive co-learning model. 

From the sounding object visual grounding and audio-visual sound separation results, we can conclude that our cyclic co-learning framework can make the two tasks benefit from each other and significantly improve both visual grounding and sound separation performance.  

\begin{table}
\centering
\scalebox{0.7}{
  \begin{tabular}{l|cc|ccc|c}
    \toprule
    Methods    & SoP~\cite{zhao2018sound} & CoSep~\cite{gao2019co} &Random Obj &CoL &CCoL & GT \\
    \midrule
    SDR   & -11.35 & -15.11 &-11.34 &14.78 &\underline{\textbf{91.07}} & 264.44 \\
     SIR  & -10.40& -12.81 &-9.09 &15.68 &\underline{\textbf{82.82}} & 260.35\\
    \bottomrule
  \end{tabular}
  }
  \caption{Audio-visual sound separation performance (with silent objects). To help readers better appreciate the results, we include SDR and SIR from ground truth sounds.}
  \label{tbl:sep_silent}
\vspace{-3mm}
\end{table}

\begin{figure}
    \centering
    \includegraphics[width=\linewidth]{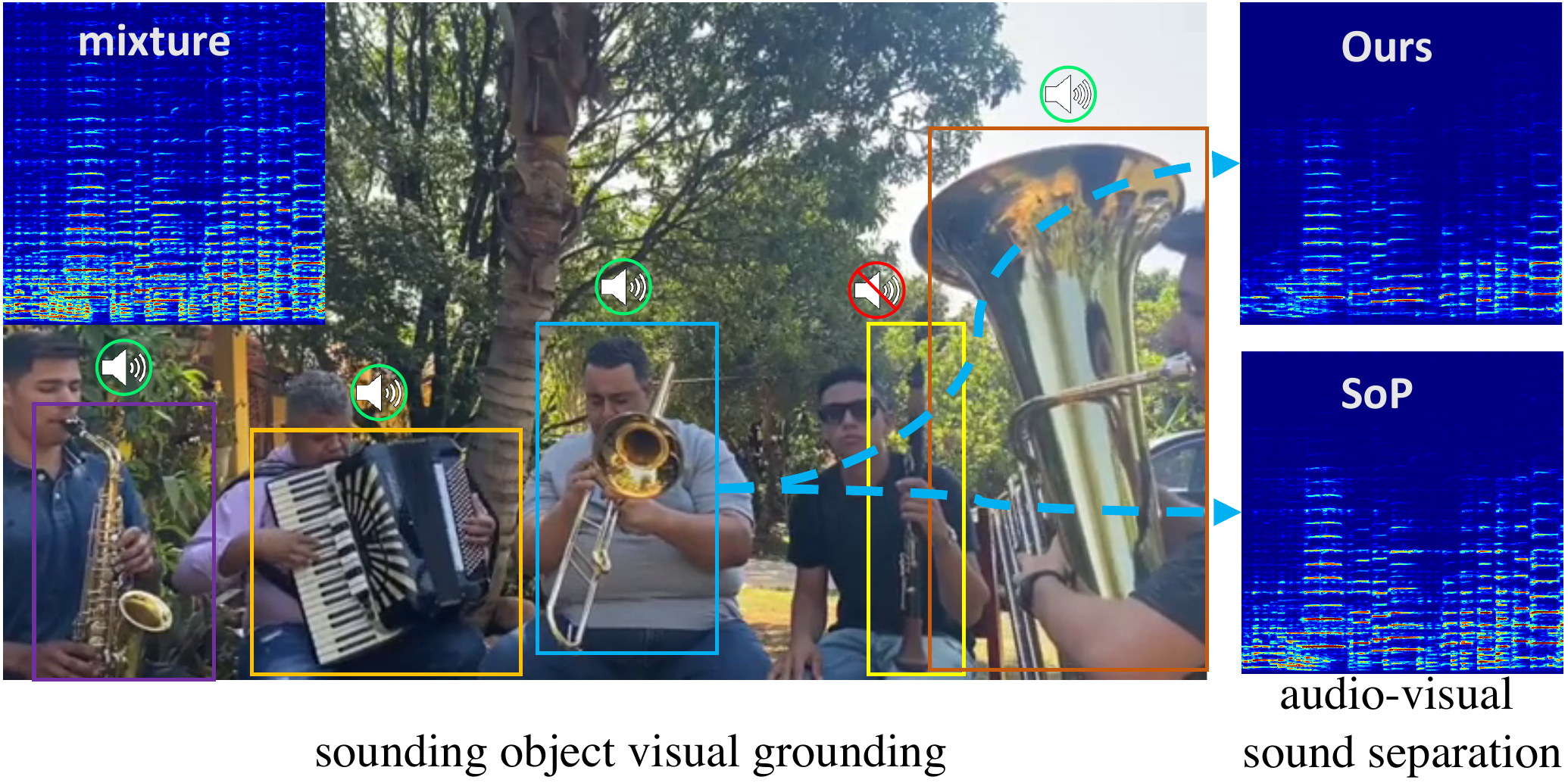}
    \caption{Real-world sounding object visual grounding and audio-visual sound separation for a challenging outdoor video with a 4-sound mixture and five instruments. From left to right, the instruments are \textit{saxophone}, \textit{accordion}, \textit{trombone}, \textit{clarinet} and \textit{tuba}. Our grounding network can successfully find audible and silent objects, and our separation network can separate the \textit{trombone} music, meanwhile suppressing unrelated sounds from other playing instruments. However, the separated sound by SoP contains much stronger noisy sounds from the \textit{accordion} and \textit{tuba}.  Note that the \textit{trombone} is an unseen category, and we have no 4-sound mixture during training. The example can demonstrate the generalization capacity of our method.}
    \label{fig:real}
    \vspace{-3mm}
\end{figure}

\subsection{Real-World Grounding and Separation}
Besides synthetic data, our grounding and separation models can also handle real-world videos, in which multiple audible and silent objects might exist in single video frames. 
The example shown in Fig.~\ref{fig:joint_loc_sep} consists of three different instruments: \textit{guitar}, \textit{flute}, and \textit{violin} in the video, in which \textit{guitar} and \textit{violin} are playing and making sounds. We can see that our grounding model can successfully identify the sounding objects: \textit{guitar} and \textit{violin} and silent object: \textit{flute}. Meanwhile, our sounding object-aware separation model can separate individual sounds for each music instrument from the sound mixture. However, the SoP using the same noisy training data as our method obtains poor separation results because it is limited in associating objects with corresponding sounds. 

Another more challenging example is illustrated in Fig.~\ref{fig:real}. There are five different instruments in the video, and four of them are playing together. Although our grounding and separation networks are not trained with 4-sound mixtures, our sounding object visual grounding network can accurately find audible and silent objects in the video, and our audio-visual separation network can separate the \textit{trombone} sound and is more capable of suppressing unrelated sounds from other three playing instruments than the SoP.   

From the above results, we learn that our separation and grounding models trained using the synthetic data can be generalized to handle real-world challenging videos. In addition, the results further demonstrate that our models can effectively identify audible and silent objects and automatically discover the association between objects and individual sounds without relying on human annotations.  

\vspace{-2mm}
\section{Limitation and Discussion}
Our sounding object visual grounding model first processes video frames and obtains object candidates by a Faster R-CNN~\cite{ren2015faster} object detector trained on a subset of Open Images dataset~\cite{krasin2017openimages}. As stated in Sec.~\ref{sec:sovg}, such a strategy of using visual object proposal has been widely employed in image captioning~\cite{karpathy2015deep,anderson2018bottom} and recent work on audio-visual learning~\cite{gao2019co}. 
Intuitively, similar to these previous works~\cite{karpathy2015deep,anderson2018bottom,gao2019co}, the grounding model performance also highly relies on the quality of the object detector. If a sounding object is not detected by the detector, our model will fail to ground it. Thus, an accurate and effective object detector is very important for the grounding model. At present, one possible way to address this limitation could be performing dense proposal of object candidates to prevent the above problem to some extent.

\vspace{-2mm}
\section{Conclusion and Future Work}
\vspace{-2mm}
\label{con}

In this paper, we introduce a cyclic co-learning framework that can jointly learn sounding object visual grounding and audio-visual sound separation. With the help of sounding object visual grounding, we propose to perform sounding object-aware sound separation to improve audio-visual sound separation. To further facilitate sounding object visual grounding learning, we use the separation model to help training sampling mining, which makes the learning process of the two tasks in a cycle and can simultaneously enhance both grounding and separation performance.
Extensive experiments can validate that the two different problems are highly coherent, and they can benefit from each other with our cyclic co-learning, and the proposed model can achieve noticeable performance on both sounding object visual grounding and audio-visual sound separation. 

There is various audio and visual content with different modality compositions in real-world videos. Besides the silent object issue, there are other challenging cases that affect audio-visual learning, as discussed below. 

There might be multiple instances of the same instrument in a video. Our separation model mainly uses encoded visual semantic information to separate sounds; however, the visual dynamics that can provide additional discriminative information to identify different instances for the same instrument are not exploited. In the future, we will consider how to incorporate dynamic visual information to further strengthen our model's ability on separating multiple sounds of the same types as in~\cite{zhao2019sound,gan2020music}. 

Sound sources are not always visible in videos. For example, guitar sound might only be background music. In this case, there are no corresponding visual objects that can be used as conditions to separate the guitar sound for existing audio-visual sound separation methods. To address this problem, we might need to first parse input sound mixtures and video frames to recognize invisible sounds and then use other reliable conditions (\eg, retrieved video frames or other semantic information) to separate the sounds.

\section{Acknowledgement} 
Y. Tian and C. Xu were supported by NSF 1741472, 1813709, and 1909912. D. Hu was supported by Fundamental Research Funds for the Central Universities, the Research Funds of Renmin University of China (NO. 21XNLG17), the Beijing Outstanding Young Scientist Program NO. BJJWZYJH012019100020098 and the 2021 Tencent AI Lab Rhino-Bird Focused Research Program (NO. JR202141). The article solely reflects the opinions and conclusions of its authors but not the funding agents.

{\small
\bibliographystyle{ieee_fullname}
\bibliography{egbib}

\begin{thebibliography}{10}\itemsep=-1pt

\bibitem{afouras2020self}
Triantafyllos Afouras, Andrew Owens, Joon~Son Chung, and Andrew Zisserman.
\newblock Self-supervised learning of audio-visual objects from video.
\newblock In {\em ECCV}, 2020.

\bibitem{anderson2018bottom}
Peter Anderson, Xiaodong He, Chris Buehler, Damien Teney, Mark Johnson, Stephen
  Gould, and Lei Zhang.
\newblock Bottom-up and top-down attention for image captioning and visual
  question answering.
\newblock In {\em Proceedings of the IEEE conference on computer vision and
  pattern recognition}, pages 6077--6086, 2018.

\bibitem{arandjelovic2018objects}
Relja Arandjelovic and Andrew Zisserman.
\newblock Objects that sound.
\newblock In {\em ECCV}, 2018.

\bibitem{bengio2009curriculum}
Yoshua Bengio, J{\'e}r{\^o}me Louradour, Ronan Collobert, and Jason Weston.
\newblock Curriculum learning.
\newblock In {\em Proceedings of the 26th annual international conference on
  machine learning}, pages 41--48, 2009.

\bibitem{bulkin2006seeing}
David~A Bulkin and Jennifer~M Groh.
\newblock Seeing sounds: visual and auditory interactions in the brain.
\newblock {\em Current opinion in neurobiology}, 16(4):415--419, 2006.

\bibitem{chen2018lip}
Lele Chen, Zhiheng Li, Ross~K Maddox, Zhiyao Duan, and Chenliang Xu.
\newblock Lip movements generation at a glance.
\newblock In {\em Proceedings of the European Conference on Computer Vision
  (ECCV)}, pages 520--535, 2018.

\bibitem{chen2019hierarchical}
Lele Chen, Ross~K Maddox, Zhiyao Duan, and Chenliang Xu.
\newblock Hierarchical cross-modal talking face generation with dynamic
  pixel-wise loss.
\newblock In {\em Proceedings of the IEEE/CVF Conference on Computer Vision and
  Pattern Recognition}, pages 7832--7841, 2019.

\bibitem{ephrat2018looking}
Ariel Ephrat, Inbar Mosseri, Oran Lang, Tali Dekel, Kevin Wilson, Avinatan
  Hassidim, William~T Freeman, and Michael Rubinstein.
\newblock Looking to listen at the cocktail party: A speaker-independent
  audio-visual model for speech separation.
\newblock {\em TOG}, 2018.

\bibitem{gan2020music}
Chuang Gan, Deng Huang, Hang Zhao, Joshua~B Tenenbaum, and Antonio Torralba.
\newblock Music gesture for visual sound separation.
\newblock In {\em Proceedings of the IEEE/CVF Conference on Computer Vision and
  Pattern Recognition}, pages 10478--10487, 2020.

\bibitem{gan2019self}
Chuang Gan, Hang Zhao, Peihao Chen, David Cox, and Antonio Torralba.
\newblock Self-supervised moving vehicle tracking with stereo sound.
\newblock In {\em Proceedings of the IEEE International Conference on Computer
  Vision}, pages 7053--7062, 2019.

\bibitem{gao2018learning}
Ruohan Gao, Rogerio Feris, and Kristen Grauman.
\newblock Learning to separate object sounds by watching unlabeled video.
\newblock In {\em Proceedings of the European Conference on Computer Vision
  (ECCV)}, pages 35--53, 2018.

\bibitem{gao20192}
Ruohan Gao and Kristen Grauman.
\newblock 2.5 d visual sound.
\newblock In {\em Proceedings of the IEEE Conference on Computer Vision and
  Pattern Recognition}, pages 324--333, 2019.

\bibitem{gao2019co}
Ruohan Gao and Kristen Grauman.
\newblock Co-separating sounds of visual objects.
\newblock In {\em ICCV}, 2019.

\bibitem{gao2019listentolook}
Ruohan Gao, Tae-Hyun Oh, Kristen Grauman, and Lorenzo Torresani.
\newblock Listen to look: Action recognition by previewing audio.
\newblock {\em arXiv preprint arXiv:1912.04487}, 2019.

\bibitem{he2016deep}
Kaiming He, Xiangyu Zhang, Shaoqing Ren, and Jian Sun.
\newblock Deep residual learning for image recognition.
\newblock In {\em CVPR}, pages 770--778, 2016.

\bibitem{hershey2016deep}
John~R Hershey, Zhuo Chen, Jonathan Le~Roux, and Shinji Watanabe.
\newblock Deep clustering: Discriminative embeddings for segmentation and
  separation.
\newblock In {\em ICASSP}, pages 31--35. IEEE, 2016.

\bibitem{hershey2000audio}
John~R Hershey and Javier~R Movellan.
\newblock Audio vision: Using audio-visual synchrony to locate sounds.
\newblock In {\em Advances in neural information processing systems}, pages
  813--819, 2000.

\bibitem{hikosaka1988polysensory}
KAZUO Hikosaka, EIICHI Iwai, H Saito, and KEIJI Tanaka.
\newblock Polysensory properties of neurons in the anterior bank of the caudal
  superior temporal sulcus of the macaque monkey.
\newblock {\em Journal of neurophysiology}, 60(5):1615--1637, 1988.

\bibitem{Hu2020CrossTaskTF}
Di Hu, Xuhong Li, Lichao Mou, P. Jin, D. Chen, L. Jing, X. Zhu, and D. Dou.
\newblock Cross-task transfer for geotagged audiovisual aerial scene
  recognition.
\newblock In {\em ECCV}, 2020.

\bibitem{hu2019deep}
Di Hu, Feiping Nie, and Xuelong Li.
\newblock Deep multimodal clustering for unsupervised audiovisual learning.
\newblock In {\em CVPR}, 2019.

\bibitem{hu2020discriminative}
Di Hu, Rui Qian, Minyue Jiang, Xiao Tan, Shilei Wen, Errui Ding, Weiyao Lin,
  and Dejing Dou.
\newblock Discriminative sounding objects localization via self-supervised
  audiovisual matching.
\newblock In {\em Advances in Neural Information Processing Systems}, 2020.

\bibitem{hu2020curriculum}
Di Hu, Zheng Wang, Haoyi Xiong, Dong Wang, Feiping Nie, and Dejing Dou.
\newblock Curriculum audiovisual learning.
\newblock {\em arXiv preprint arXiv:2001.09414}, 2020.

\bibitem{huang2012singing}
Po-Sen Huang, Scott~Deeann Chen, Paris Smaragdis, and Mark Hasegawa-Johnson.
\newblock Singing-voice separation from monaural recordings using robust
  principal component analysis.
\newblock In {\em 2012 IEEE International Conference on Acoustics, Speech and
  Signal Processing (ICASSP)}, pages 57--60. IEEE, 2012.

\bibitem{huang2015joint}
Po-Sen Huang, Minje Kim, Mark Hasegawa-Johnson, and Paris Smaragdis.
\newblock Joint optimization of masks and deep recurrent neural networks for
  monaural source separation.
\newblock {\em IEEE/ACM Transactions on Audio, Speech, and Language
  Processing}, 23(12):2136--2147, 2015.

\bibitem{karpathy2015deep}
Andrej Karpathy and Li Fei-Fei.
\newblock Deep visual-semantic alignments for generating image descriptions.
\newblock In {\em Proceedings of the IEEE conference on computer vision and
  pattern recognition}, pages 3128--3137, 2015.

\bibitem{kazakos2019epic}
Evangelos Kazakos, Arsha Nagrani, Andrew Zisserman, and Dima Damen.
\newblock Epic-fusion: Audio-visual temporal binding for egocentric action
  recognition.
\newblock In {\em Proceedings of the IEEE International Conference on Computer
  Vision}, pages 5492--5501, 2019.

\bibitem{kidron2005pixels}
Einat Kidron, Yoav~Y Schechner, and Michael Elad.
\newblock Pixels that sound.
\newblock In {\em 2005 IEEE Computer Society Conference on Computer Vision and
  Pattern Recognition (CVPR'05)}, volume~1, pages 88--95. IEEE, 2005.

\bibitem{kingma2014adam}
Diederik~P Kingma and Jimmy Ba.
\newblock Adam: A method for stochastic optimization.
\newblock {\em arXiv preprint arXiv:1412.6980}, 2014.

\bibitem{korbar2019scsampler}
Bruno Korbar, Du Tran, and Lorenzo Torresani.
\newblock Scsampler: Sampling salient clips from video for efficient action
  recognition.
\newblock In {\em Proceedings of the IEEE International Conference on Computer
  Vision}, pages 6232--6242, 2019.

\bibitem{krasin2017openimages}
Ivan Krasin, Tom Duerig, Neil Alldrin, Vittorio Ferrari, Sami Abu-El-Haija,
  Alina Kuznetsova, Hassan Rom, Jasper Uijlings, Stefan Popov, Andreas Veit,
  et~al.
\newblock Openimages: A public dataset for large-scale multi-label and
  multi-class image classification.
\newblock {\em Dataset available from https://github. com/openimages}, 2017.

\bibitem{lin2019dual}
Yan-Bo Lin, Yu-Jhe Li, and Yu-Chiang~Frank Wang.
\newblock Dual-modality seq2seq network for audio-visual event localization.
\newblock In {\em ICASSP 2019-2019 IEEE International Conference on Acoustics,
  Speech and Signal Processing (ICASSP)}, pages 2002--2006. IEEE, 2019.

\bibitem{owens2018audio}
Andrew Owens and Alexei~A Efros.
\newblock Audio-visual scene analysis with self-supervised multisensory
  features.
\newblock {\em European Conference on Computer Vision (ECCV)}, 2018.

\bibitem{paszke2019pytorch}
Adam Paszke, Sam Gross, Francisco Massa, Adam Lerer, James Bradbury, Gregory
  Chanan, Trevor Killeen, Zeming Lin, Natalia Gimelshein, Luca Antiga, et~al.
\newblock Pytorch: An imperative style, high-performance deep learning library.
\newblock In {\em Advances in Neural Information Processing Systems}, pages
  8024--8035, 2019.

\bibitem{qian2020multiple}
Rui Qian, Di Hu, Heinrich Dinkel, Mengyue Wu, Ning Xu, and Weiyao Lin.
\newblock Multiple sound sources localization from coarse to fine.
\newblock In {\em ECCV}, 2020.

\bibitem{raffel2014mir_eval}
Colin Raffel, Brian McFee, Eric~J Humphrey, Justin Salamon, Oriol Nieto, Dawen
  Liang, Daniel~PW Ellis, and C~Colin Raffel.
\newblock mir\_eval: A transparent implementation of common mir metrics.
\newblock In {\em ISMIR}, 2014.

\bibitem{rahman2019watch}
Tanzila Rahman, Bicheng Xu, and Leonid Sigal.
\newblock Watch, listen and tell: Multi-modal weakly supervised dense event
  captioning.
\newblock In {\em Proceedings of the IEEE International Conference on Computer
  Vision}, pages 8908--8917, 2019.

\bibitem{ren2015faster}
Shaoqing Ren, Kaiming He, Ross Girshick, and Jian Sun.
\newblock Faster r-cnn: Towards real-time object detection with region proposal
  networks.
\newblock In {\em NIPS}, 2015.

\bibitem{ronneberger2015u}
Olaf Ronneberger, Philipp Fischer, and Thomas Brox.
\newblock U-net: Convolutional networks for biomedical image segmentation.
\newblock In {\em International Conference on Medical image computing and
  computer-assisted intervention}, 2015.

\bibitem{rouditchenko2019self}
Andrew Rouditchenko, Hang Zhao, Chuang Gan, Josh McDermott, and Antonio
  Torralba.
\newblock Self-supervised segmentation and source separation on videos.
\newblock In {\em Proceedings of the IEEE Conference on Computer Vision and
  Pattern Recognition Workshops}, pages 0--0, 2019.

\bibitem{senocak2018learning}
Arda Senocak, Tae-Hyun Oh, Junsik Kim, Ming-Hsuan Yang, and In So~Kweon.
\newblock Learning to localize sound source in visual scenes.
\newblock In {\em Proceedings of the IEEE Conference on Computer Vision and
  Pattern Recognition}, pages 4358--4366, 2018.

\bibitem{simonyan2014very}
Karen Simonyan and Andrew Zisserman.
\newblock Very deep convolutional networks for large-scale image recognition.
\newblock {\em ICLR}, 2014.

\bibitem{stein1993merging}
Barry~E Stein and M~Alex Meredith.
\newblock {\em The merging of the senses.}
\newblock The MIT Press, 1993.

\bibitem{Tian_2019_CVPR_Workshops}
Yapeng Tian, Chenxiao Guan, Goodman Justin, Marc Moore, and Chenliang Xu.
\newblock Audio-visual interpretable and controllable video captioning.
\newblock In {\em The IEEE Conference on Computer Vision and Pattern
  Recognition (CVPR) Workshops}, June 2019.

\bibitem{tian2020avvp}
Yapeng Tian, Dingzeyu Li, and Chenliang Xu.
\newblock Unified multisensory perception: Weakly-supervised audio-visual video
  parsing.
\newblock In {\em ECCV}, 2020.

\bibitem{tian2018audio}
Yapeng Tian, Jing Shi, Bochen Li, Zhiyao Duan, and Chenliang Xu.
\newblock Audio-visual event localization in unconstrained videos.
\newblock In {\em Proceedings of the European Conference on Computer Vision
  (ECCV)}, pages 247--263, 2018.

\bibitem{wang2018watch}
Xin Wang, Yuan-Fang Wang, and William~Yang Wang.
\newblock Watch, listen, and describe: Globally and locally aligned cross-modal
  attentions for video captioning.
\newblock {\em arXiv preprint arXiv:1804.05448}, 2018.

\bibitem{wu2019DAM}
Yu Wu, Linchao Zhu, Yan Yan, and Yi Yang.
\newblock Dual attention matching for audio-visual event localization.
\newblock In {\em Proceedings of the IEEE International Conference on Computer
  Vision (ICCV)}, 2019.

\bibitem{xu2019recursive}
Xudong Xu, Bo Dai, and Dahua Lin.
\newblock Recursive visual sound separation using minus-plus net.
\newblock In {\em Proceedings of the IEEE International Conference on Computer
  Vision}, pages 882--891, 2019.

\bibitem{Xu2021visual}
Xudong Xu, Hang Zhou, Ziwei Liu, Bo Dai, Xiaogang Wang, and Dahua Lin.
\newblock Visually informed binaural audio generation without binaural audios.
\newblock In {\em Proceedings of the IEEE/CVF Conference on Computer Vision and
  Pattern Recognition}, 2021.

\bibitem{zhao2019sound}
Hang Zhao, Chuang Gan, Wei-Chiu Ma, and Antonio Torralba.
\newblock The sound of motions.
\newblock In {\em Proceedings of the IEEE International Conference on Computer
  Vision}, pages 1735--1744, 2019.

\bibitem{zhao2018sound}
Hang Zhao, Chuang Gan, Andrew Rouditchenko, Carl Vondrick, Josh McDermott, and
  Antonio Torralba.
\newblock The sound of pixels.
\newblock In {\em ECCV}, 2018.

\bibitem{Zhou2021pose}
Hang Zhou, Yasheng Sun, Wu Wayne, Chen~Change Loy, Xiaogang Wang, and Liu
  Ziwei.
\newblock Pose-controllable talking face generation by implicitly modularized
  audio-visual representation.
\newblock In {\em Proceedings of the IEEE/CVF Conference on Computer Vision and
  Pattern Recognition}, 2021.

\bibitem{zhou2020sep}
Hang Zhou, Xudong Xu, Dahua Lin, Xiaogang Wang, and Ziwei Liu.
\newblock Sep-stereo: Visually guided stereophonic audio generation by
  associating source separation.
\newblock In {\em European Conference on Computer Vision}, pages 52--69.
  Springer, 2020.

\end{thebibliography}
}
\newpage
\appendix

\definecolor{Gray}{gray}{0.9}
\definecolor{LightCyan}{rgb}{0.88,1,1}

\section*{Appendix}

We include this appendix to describe more details about audio-visual sound separation network and our implementation. Moreover, we provide more evaluation results for silent sound separation.

\section*{Audio-Visual Sound Separation Network}
\label{sec:avssn}
We adopt the same audio-visual separator as in~\cite{zhao2018sound}, which consists of three modules: audio network, visual network, and audio-visual sound synthesizing network. 

We use Time-Frequency representation of sound and project raw waveform to spectrogram with the STFT. The audio network transforms magnitude of STFT spectrogram $S_m$ from the input audio mixture to a $C$-channel feature map $A_m\in \mathcal{R}^{C\times F\times T}$ with an U-Net~\cite{ronneberger2015u} structure, where $C = 32$ in our experiments. 
We use an ResNet-18~\cite{he2016deep} followed by a linear layer to predict a $C$-dimension object feature $f_{o_i^{(k)}}\in \mathcal{R}^{C}$ for each object $O_i^{(k)}$. The audio-visual sound synthesizing network takes $A_m$ and $f_{o_i^{(k)}}$ as inputs and output a spectrogram mask $M_i^{(k)}\in \mathcal{R}^{F\times T}$ with a linearly transformed dot product of the audio and visual features. The separated sound spectrogram: $S_i^{(k)} = S_m \odot M_i^{(k)}$ can be obtained by masking the sound mixture, where $\odot$ is the element-wise multiplication operator.
The waveform $s_i^{(k)}$ of the object can be reconstructed by the inverse short-time Fourier transform. 

\section*{Implementation Details}

We train our networks using PyTorch~\cite{paszke2019pytorch} library with 4 NVIDIA 1080Ti GPUs. The batch size and epoch number are set to 48 and 60, respectively. The learning rate is set to $1e-4$ and it will decrease by multiplying $0.1$ at $30$- and $50$-th epochs, respectively. For the three-step training, we train 60 epochs for each step. We conduct ablation studies with several models: Grounding only, Random Obj, CoL, and CCoL, they are defined as follows:

\noindent \textbf{Grounding Only:} The Grounding only model is trained only with grounding losses: $l_{grd_s}$;

\noindent \textbf{Random Obj:} This baseline randomly selects objects from detected object proposals to train a audio-visual sound separation model;

\noindent \textbf{CoL:} The co-learning model (CoL) jointly learns visual grounding and sound separation using the $l_{col}$;

\noindent \textbf{CCoL:} The cyclic co-learning (CCoL) further strengthens the interaction between the two tasks optimized by $l_{ccol}$. 

\section*{Additional Evaluation}
To further validate silent sound separation performance, we compute the sound energy for separated sounds of silent objects. We empirically set a threshold: 20 to decide whether the separation is successful. The success rates for silent objects classification are 8.6\% and 92.3\% for SoP and CCoL, respectively. An interesting observation is that the success rate of our CCoL is even higher than the corresponding grounding accuracy, which demonstrates that our model tends to generate weak sounds for silent objects.

\end{document}